\documentclass[conference]{IEEEtran}
\IEEEoverridecommandlockouts
\usepackage{cite}
\usepackage{ragged2e}
\usepackage{amsmath,amssymb,amsfonts}
\usepackage{algorithmic}
\usepackage{graphicx}
\usepackage{textcomp}
\usepackage{multirow}
\usepackage[most]{tcolorbox} 
\usepackage{xcolor}
\usepackage[font=footnotesize, labelfont=footnotesize]{caption}
\usepackage{subcaption}
\def\BibTeX{{\rm B\kern-.05em{\sc i\kern-.025em b}\kern-.08em
    T\kern-.1667em\lower.7ex\hbox{E}\kern-.125emX}}
\begin{document}

\newtcolorbox{llmbox}[1][]{
  enhanced,
  breakable,                 
  arc=4pt, 
  colback=blue!3,            
  colframe=blue!60!black,    
  coltitle=blue!20!black,
  coltext=red!50!black, 
  fontupper=\small\justifying, 
  fonttitle=\bfseries\ttfamily,
  left=1.2ex, right=1.2ex, top=0.8ex, bottom=1ex,
  boxsep=0.6ex,
  title=LLM Output,          
  #1                         
}

\title{Causal Autoencoder-like Generation of Feedback Fuzzy Cognitive Maps with an LLM Agent}

\author{\IEEEauthorblockN{\ Akash Kumar Panda}
\IEEEauthorblockA{\textit{\hspace{-0.2in}Department of Electrical and  Computer Engineering} \\
\textit{University of Southern California}\\
Los Angeles, USA \\
akashpan@usc.edu }
\and
\IEEEauthorblockN{\hspace{0.5in} Olaoluwa Adigun}
\IEEEauthorblockA{\textit{\hspace{0.5in} School of Computing and Information Sciences} \\
\textit{\hspace{0.5in} Florida International University}\\
\hspace{0.5in} Miami, USA \\
\hspace{0.5in} olaadigu@fiu.edu}
\and
\IEEEauthorblockN{\hspace{0.0in}}
\IEEEauthorblockA{\textit{} \\
\textit{}\\
}
\and
\IEEEauthorblockN{\hspace{1.9in}Bart Kosko}
\IEEEauthorblockA{\textit{\hspace{1.9in}Department of Electrical and  Computer Engineering} \\
\textit{\hspace{1.9in}University of Southern California}\\
\hspace{1.9in} Los Angeles, USA \\
\hspace{1.9in} kosko@usc.edu}
}

\maketitle

\begin{abstract}
A large language model (LLM) can map a feedback causal fuzzy cognitive map (FCM) into text and then reconstruct the FCM from the text.
This explainable AI system approximates an identity map from the FCM to itself and resembles the operation of an autoencoder (AE).
Both the encoder and the decoder explain their decisions in contrast to black-box AEs. 
Humans can read and interpret the encoded text in contrast to the hidden variables and synaptic webs in AEs. 
The LLM agent approximates the identity map through a sequence of system instructions that does not compare the output to the input. 
The reconstruction is lossy because it removes weak causal edges or rules while it preserves strong causal edges. 
The encoder preserves the strong causal edges even when it trades off some details about the FCM to make the text sound more natural. 
\end{abstract}


\section{Generating Causal Feedback Fuzzy Cognitive Maps by Identity Approximations}\label{Introduction}

How do we generate representative text from a causal feedback semantic network?  
This task is at least as difficult as the reverse problem of how do we reliably map text to a causal feedback semantic network such as a feedback fuzzy cognitive map (FCM)\cite{kosko1986fuzzy,kosko1988hidden,osoba2017fuzzy,ziv2018potential,glykas2010fuzzy,papageorgiou2013fuzzy,stach2010divide,taber2007quantization} dynamical system.  

We harness large language models (LLMs) to approximate an identity map  $\Phi:\mathcal{F}\rightarrow\mathcal{F}$  from the feedback causal map $F\in\mathcal{F}$ to $\Phi(F)\approx F$ by way of representative text.  
The mapping structure  $\Phi$  resembles the forward-backward structure of autoencoders (AEs) often found in AI image generators and LLMs.  
But those semi-supervised maps rely on black-box neural networks with simple feedforward layers of neurons and that have no dynamics.  
FCM knowledge graphs are explainable AI (XAI) models that list their local causal rules in their edge matrix.  
Their feedback structure defines a dynamical system with rich output equilibria that can act as answers to user questions and can guide agentic-like tasks.

\begin{figure*}[htbp]
\centerline{\includegraphics[width=0.9\textwidth]{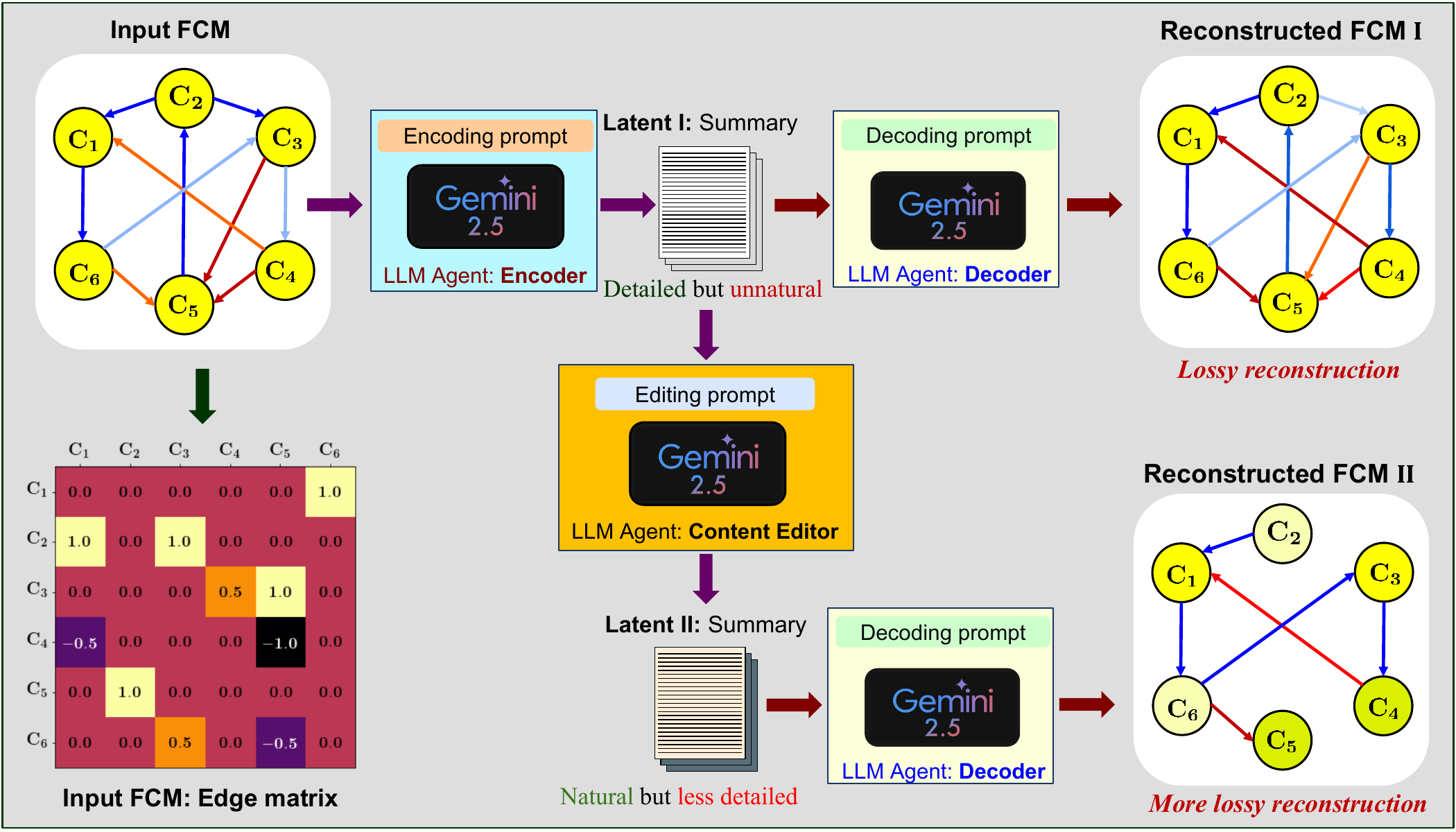}}
\caption{{Autoencoding Fuzzy Cognitive Maps (FCMs) with a single LLM agent and multi-prompting: 
The input FCM is on the top-left. 
Its edge matrix $E$ is on the bottom-left. 
The edge matrix colors correspond to the edge weights. 
Higher edge weights correspond to brighter colors.
The LLM agent with the encoding prompt converts the input FCM into the text description latent I. 
This text description is a detailed description of the input but sounds unnatural. 
The LLM agent with the content-editing prompt reworks latent I into latent II. 
The result sounds more natural but sacrifices some detail. 
The LLM agent with the decoding prompt reconstructs FCMs from their text description in latent I and latent II. 
The top-right FCM shows that unnatural yet detailed latent I gives a lossy FCM reconstruction. 
The less detailed yet natural sounding latent II gives a lossier reconstructed FCM in the bottom-right FCM.}
}
\vspace{-0.10in}
\label{fig:FCM_AE_architecture}
\end{figure*}

Figure~\ref{fig:FCM_AE_architecture} shows this autoencoder-like identity approximator.
The LLM agent with the encoding prompt maps from the FCM to text.
The LLM agent with the decoding prompts then maps the text back to the reconstructed FCM. 
The encoded text summary latent I sounds unnatural when the encoder focuses on capturing every detail of the FCM. 
The LLM agent with the content editing prompt can rewrite the text to make it sound more natural. 
But this might come with a loss in detail leading to a lossy reconstruction. 

Consider an FCM that models the cause of clinical depression \cite{billis2014decision}. 
Column 2 of Table~\ref{tab:depression_fcm} gives the nodes of this FCM. 
The ${7}^{\text{th}}$ node $C_7$ represents ``fatigue or loss of energy" and the ${9}^{\text{th}}$ node $C_9$ represents ``loss of appetite". 
Figure~\ref{fig:depression_target} shows the edge matrix $E$ of this FCM. 
It also shows that there is a causal edge $e_{97}$ ``loss of appetite" $\rightarrow$ ``fatigue or loss of energy" with weight $w_{97} = 0.8$.  
The LLM agent with the encoding prompt translated this edge to this sentence in the latent I summary:
\begin{llmbox}[title={}, colback=yellow!20, colframe=gray!60]
\noindent {{\textbf{`Loss of appetite'}} strongly causes {\textbf{`fatigue or loss of energy'}} and significantly increases {\textbf{`psychomotor retardation'}} and {\textbf{`reduced interest for daily functioning'}.}}
\end{llmbox}
\noindent The LLM agent decodes this summary as follows: $C_7$ is ``fatigue or loss of energy", $C_9$ is ``loss of appetite", and the edge $e_{97}$ ``loss of appetite" $\rightarrow$ ``fatigue or loss of energy" has weight $w_{97} = 0.8$.
The agent was too focused on capturing the exact nodes and edges so the text sounds unnatural. 

The LLM agent with the content-editing prompt rewrites the text summary latent I to make it sound more natural.
This gives the natural sounding text summary latent II with the corresponding sentence:
\begin{llmbox}[title={}, colback=yellow!20, colframe=gray!60]
\noindent {Even a loss of  {\textbf{appetite}} contributes to the cycle by strongly causing {\textbf{fatigue}} and significantly increasing {\textbf{psychomotor retardation}} and a loss of {\textbf{interest in daily activities}}.}
\end{llmbox}
\noindent  This sentence from the latent II summary then translates to the following: $C_7$ is ``fatigue or loss of energy", $C_9$ is ``appetite", and the fuzzy or partial causal edge $e_{97}$ ``appetite" $\rightarrow$ ``fatigue or loss of energy" has weight $w_{97} = -0.8$.
Note that the source node is ``appetite" instead of ``loss of appetite".

Section~\ref{FCM} explains how feedback FCMs work.
It shows how FCMs model causal dynamical systems as weighted directed graphs and also explains what their nodes and edges represent. 
This section also explains how FCMs evolve in discrete time through matrix-multiplication and nonlinear operations to give FCM limit-cycle equilibria. 
Section~\ref{FCM} concludes by showing how to mix FCMs into a new FCM.
FCMs allow knowledge combination through convex mixing unlike directed acyclic graphs (DAGs) \cite{neuberg2003causality} or Markov chains with different states.

Section~\ref{AE} discusses autoencoding and its variants including ordinary and variational autoencoder networks. 
This section also explains how our FCM ``autoencoder" differs from the usual autoencoder networks in terms of explainability and supervision. 
Section~\ref{LLM} briefly explains LLMs and their single-agent and multi-prompting version. 
It shows how system instructions can manipulate the behavior of the LLM agent. 
This section also explains how the LLM's Natural Language Processing (NLP) and Named Entity Recognition (NER) capabilities \cite{wang2023gpt} encode and reconstruct FCMs. 

Section~\ref{Approach} explains how we use LLMs to map FCMs to text and then back to FCMs. 
It goes through every system instruction that multi-prompts LLM agent to systematically convert an FCM to text. 
It also goes through the system instructions that help the LLM reconstruct an FCM from its text description. 

Section~\ref{Experiments} discusses our experiments and their results. 
Google's Gemini 2.5 Pro\cite{Gemini2.5ProReport} takes 3 different FCMs as input and then converts them to their corresponding text descriptions. 
The Gemini LLM then reconstructs the FCMs from those text descriptions. 
The $1^{\text{st}}$ columns of Tables~\ref{tab:depression_fcm}, \ref{tab:pruned_depression}, and \ref{tab:celiac_disease}  describe the nodes on the input FCMs.
Figures \ref{fig:depression_target}, \ref{fig:pruned_depression_target}, and \ref{fig:celiac_disease_target} show their respective edges.

The $2^{\text{nd}}$ and $3^{\text{rd}}$ columns of Tables~\ref{tab:depression_fcm}, \ref{tab:pruned_depression}, and \ref{tab:celiac_disease} give the nodes of the corresponding reconstructed FCMs. 
Figures~\ref{fig:depression_prediction}$-$\ref{fig:depression_prediction_refined_positive}, \ref{fig:pruned_depression_prediction}$-$\ref{fig:pruned_depression_prediction_refined}, and \ref{fig:celiac_disease_prediction}$-$\ref{fig:celiac_disease_prediction_refined} show the edge matrices of those reconstructed FCMs. 

Section~\ref{Discussion} discusses the trade-off between natural-sounding text and FCM-reconstruction accuracy. 
It also shows that our lossy reconstruction still preserves the strong causal connections in the FCM. 
This section also explains how some nodes get flipped during FCM reconstruction.

\section{How FCMs, AEs, and LLMs Work}

\subsection{Fuzzy Cognitive Maps}\label{FCM}

FCMs model causal dynamical systems as directed weighted cyclic graphs. 
They allow fuzzy or partial causality and feedback cycles unlike causal graphs such as Bayesian belief networks (BBNs) that allow only DAGs \cite{neuberg2003causality}. 
Feedback lets FCMs model dynamical systems with complex equilibria such as limit cycles. 
FCMs with \emph{different} nodes also mix through convex combination unlike DAGs or Markov chains. 

\subsubsection{Causal model}

The FCMs model the causal variables in the dynamical system as ``concept nodes" of the directed graph. 
The nodes are associated with some kind of magnitude such that the corresponding causal variable can ``increase", ``improve", or ``intensify" because of the other causal variables of the system. 

Let $C_i$ and $C_j$ denote the respective ${i}^{\text{th}}$ and $j^{\text{th}}$ concept nodes or causal variables in the FCM. 
The edge $e_{ij}$ from $C_i$ to $C_j$ connects the nodes if ``$C_i$ causes $C_j$". 
The weight $w_{ij}\in[-1,1]$ on the edge $e_{ij}$ describes the degree to which $C_i$ causes $C_j$. 
A positive $w_{ij}$ means that an increase in $C_i$ causes $C_j$ to increase or a decrease in $C_i$ causes $C_j$ to decrease.
A negative $w_{ij}$ value means that an increase in $C_i$ causes $C_j$ to decrease or a decrease in $C_i$ causes $C_j$ to increase.
The magnitude of $w_{ij}$ shows the strength of the causal connection. 
A magnitude close to 1 denotes a ``strong" causal connection and a magnitude close to 0 denotes a ``weak" causal connection. 

The $1^{\text{st}}$ column of Table~\ref{tab:depression_fcm} and Figure~\ref{fig:depression_target} give the respective nodes $C_1$--$C_{14}$ and the edge matrix $E$ of a 14-node FCM that models causes of clinical depression\cite{billis2014decision}. 
The ${i}^{\text{th}}$ row and the ${j}^{\text{th}}$ column of the edge matrix $E$ gives the value of $w_{ij}$ for edge $e_{ij}$.

\subsubsection{Discrete time-evolution}

The $n$-dimensional row vector $C(t)\in[0,1]^n$ represents the state of a $n$-node FCM at discrete time $t$. 
The ${k}^{\text{th}}$ component $C_k(t)$ of this state vector $C(t)$ denotes the state of the ${k}^{\text{th}}$ concept node in the FCM at time $t$. 
The ${k}^{\text{th}}$ node is ``active" if $C_k(t)$ is close to 1 and it is ``inactive" if $C_k(t)$ is close to 0. 
The causal variables corresponding to the active nodes are present in the system at time $t$ and the variables corresponding to the inactive nodes are absent from the system at time $t$. 

The FCM's state vector $C(t)$ evolves in discrete time $t$ through matrix-multiplication and nonlinear squashing. 
The state of the FCM $C(t+1)$ at time $t+1$ depends on the state of the FCM $C(t)$ at time $t$ and on the edge matrix $E$: 
\begin{align}\label{fcm-update}
    C_j(t+1) = \phi\bigg(\sum_{i=1}^n  C_i(t)w_{ij}\bigg)
\end{align}
where $\phi$ is an increasing nonlinear function bounded between zero and one.

The sequence of FCM state vectors $C(0)$, $C(1)$, $C(2)$, ... describes the trajectory of the FCM in discrete time $t$ starting from the initial state $C(0)$. 
This qualitatively represents the corresponding trajectory of the dynamical system that the FCM models. 
The limiting behavior of this sequence gives the equilibrium behavior of the FCM. 
The FCM converges to a fixed point if $C(t)$ converges to a constant row-vector. 
The FCM converges to a $k$-step limit cycle if $C(t) = C(t+k)$ at some point in the trajectory. 
Then a sequence of $k$ state vectors repeats itself over and over again. 

The FCM equilibria partition the input space. 
The set of all initial states that lead to a certain equilibrium makes up the basin of that equilibrium. 
The map from the basins to the equilibrium characterizes the FCM and also the dynamical system that the FCM models. 

\subsubsection{FCM Mixing}

FCMs allow mixing through convex combination. 
Let $S_1$, $S_2$, ..., $S_m$ denote the respective node sets of $m$ FCMs. 
The node-set of the $N$-node FCM mixture is $S = S_1\cup S_2\cup...\cup S_m$. 
The $N\times N$ edge matrix $\Tilde{E}_k$ pads the ${k}^{\text{th}}$ FCM's edge matrix $E_k$ with zero-rows and zero-columns corresponding to the nodes in $S-S_k$. 
The edge matrix $E$ of the FCM mixture is
\begin{align}
    E = \sum_{k=1}^m v_k\Tilde{E_k}
\end{align}
where $v_k$ are convex mixing weights such that $v_k\geq0$ and $\sum_{k=1}^mv_k=1$. 

FCM mixing is \emph{closed}: Mixing FCMs gives back an FCM. 
This is not true in general for BBN DAGs or for the stochastic matrices of Markov chains. 
Mixed DAGs can have cycles and mixed different-state stochastic matrices may not be stochastic. 

\subsection{Autoencoders}\label{AE}
Autoencoding is a form of identity mapping from a domain back to itself.
This involves a two-step process: (1) transforming the input into a reduced form in the latent space, and (2) reconstructing the input with minimal loss. 
The first step is encoding and the second step is decoding.
Autoencoding applies to various domains including signal processing, information theory, and statistical learning.

Autoencoder networks are a family of autoencoding models that use an encoder network and a decoder network.
The encoder network maps the input data to its latent variable and the decoder network reconstructs the input data from the latent variable.
Variational autoencoders are variants of autoencoder networks that map to constrained latent variables and this makes them suitable for generative tasks\cite{kingma2013auto,higgins2017beta, burgess2018understanding,kosko2024bidirectional}.
Autoencoder networks can solve various problems including dimensionality reduction \cite{vincent2010stacked}, denoising \cite{bengio2013generalized}, anomaly detection \cite{sakurada2014anomaly}, image compression \cite{theis2017lossy}, and feature extraction \cite{gogna2019discriminative,meng2017relational}.

There are other forms of autoencoding that replace neural network models with other techniques for encoding and decoding. 
Examples of such techniques include Principal Component Analysis (PCA), wavelet transforms, and dictionary learning.

The encoder and the decoder
 networks of the autoencoder are usually black boxes and do not explain their decisions in encoding and decoding. 
The latent variables the encoder maps to is usually not human interpretable. 
Also it relies on a loss function that compares the reconstructed pattern to the input pattern. 

We present a way of multi-prompting a LLM system so that it achieves an identity map from the FCM to text and back to FCM just like an autoencoder but its system instructions can explain its decisions during the encoding and the decoding process. 
Its latent variables are also human-interpretable text descriptions of the FCM. 
It can give reasons for its decisions during reconstruction by quoting from the text that the decoder takes as input. 
Also it achieves the identity map without having to compare the reconstructed FCM to the target FCM and just by following a sequence of carefully designed system instructions. 

\subsection{Large Language Models (LLMs)}\label{LLM}

LLMs are AI systems that have trained on the vast corpora of human-generated text for the purpose of generating human-like language.
These models use the transformer neural architecture \cite{vaswani2017attention} with an attention mechanism and billions or trillions of parameters to learn human linguistic patterns.
This enables LLM to perform tasks such as text generation, summarization, translation, code-synthesis, and NER.
LLMs have become foundational in NLP and they power chatbots, virtual assistants, and enterprise automation across industries.

Single-agent multi-prompting is a strategy where a single LLM instance uses successive multiple-structured prompts to solve a complex problem.
This approach leverages the internal capabilities of the LLM agent to handle sequential or parallel subtasks using techniques such as chain-of-thought prompting, role-playing, or staged queries.
Applications of single-agent multi-prompting span education, coding, planning, and decision support where task complexity is managed through prompt design rather than architectural overhead.

\begin{figure*}[!t]
\begin{subfigure}{0.50\linewidth}
\centering
\includegraphics[width=0.9\textwidth, height=0.9\textwidth]{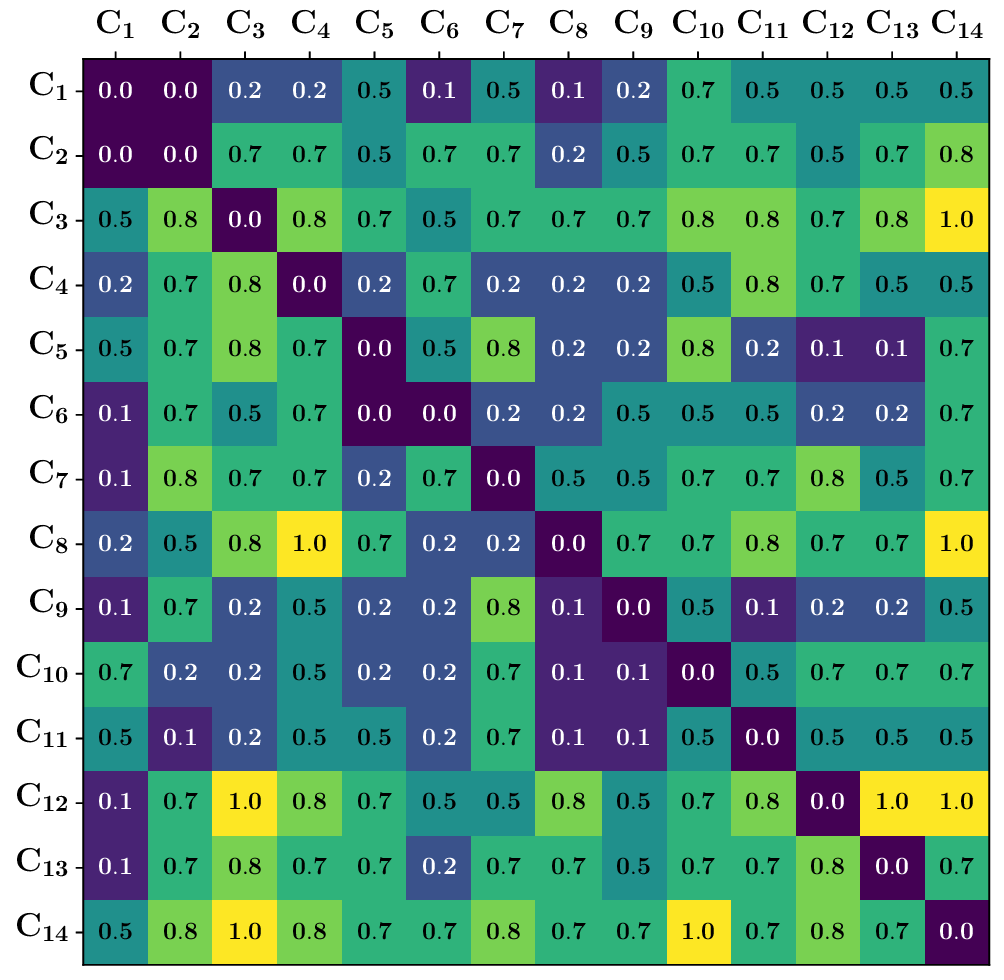}
\caption{\small{Edge matrix $E$ of the target FCM.}}
\label{fig:depression_target}
\end{subfigure}
\begin{subfigure}{0.5\linewidth}
\centering
\includegraphics[width=0.9\textwidth, height=0.9\textwidth]{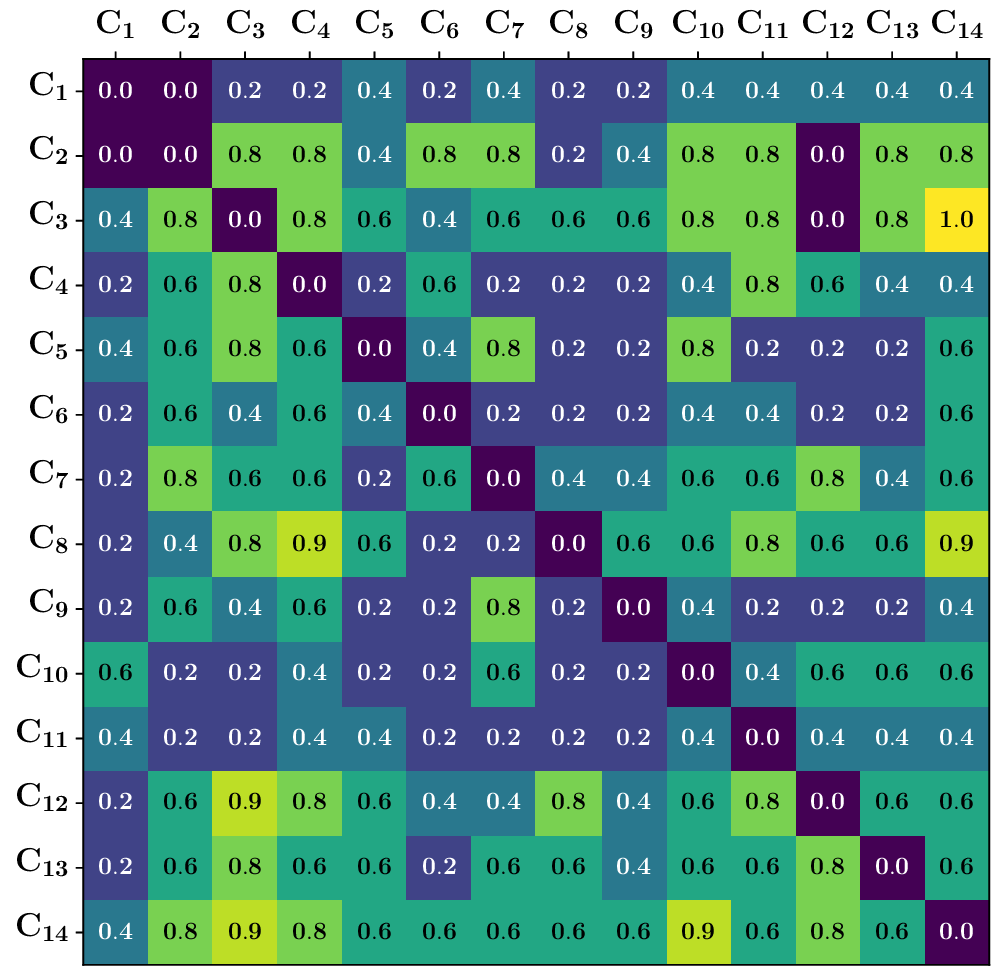} 
\caption{\small{Reconstructed edges $E_1$ from latent I. }}
\label{fig:depression_prediction}
\end{subfigure}
\begin{subfigure}{0.5\linewidth}
\centering
\vspace{0.2in}
\includegraphics[width=0.9\textwidth, height=0.9\textwidth]{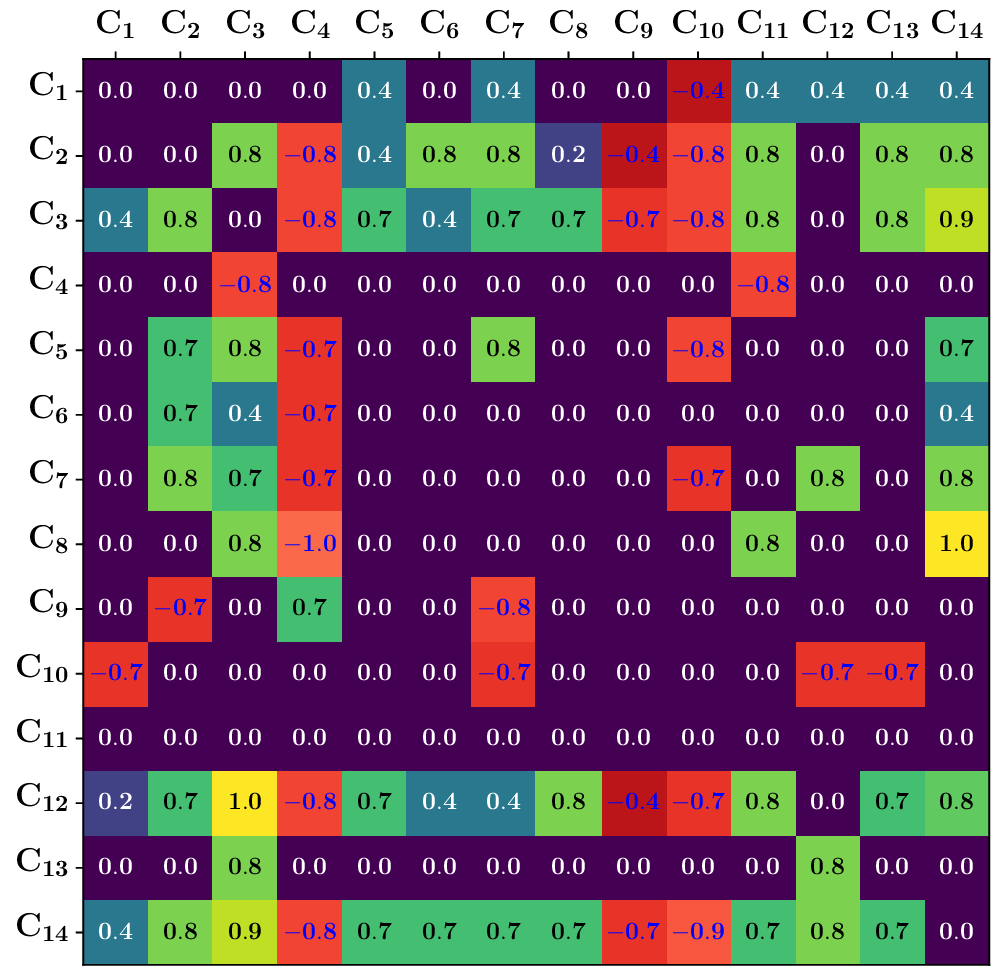}
\caption{\small{Reconstructed edges $E_2$ from latent II.}}
\label{fig:depression_prediction_refined}
\end{subfigure}
\begin{subfigure}{0.5\linewidth}
\centering
\vspace{0.2in}
\includegraphics[width=0.9\textwidth, height=0.9\textwidth]{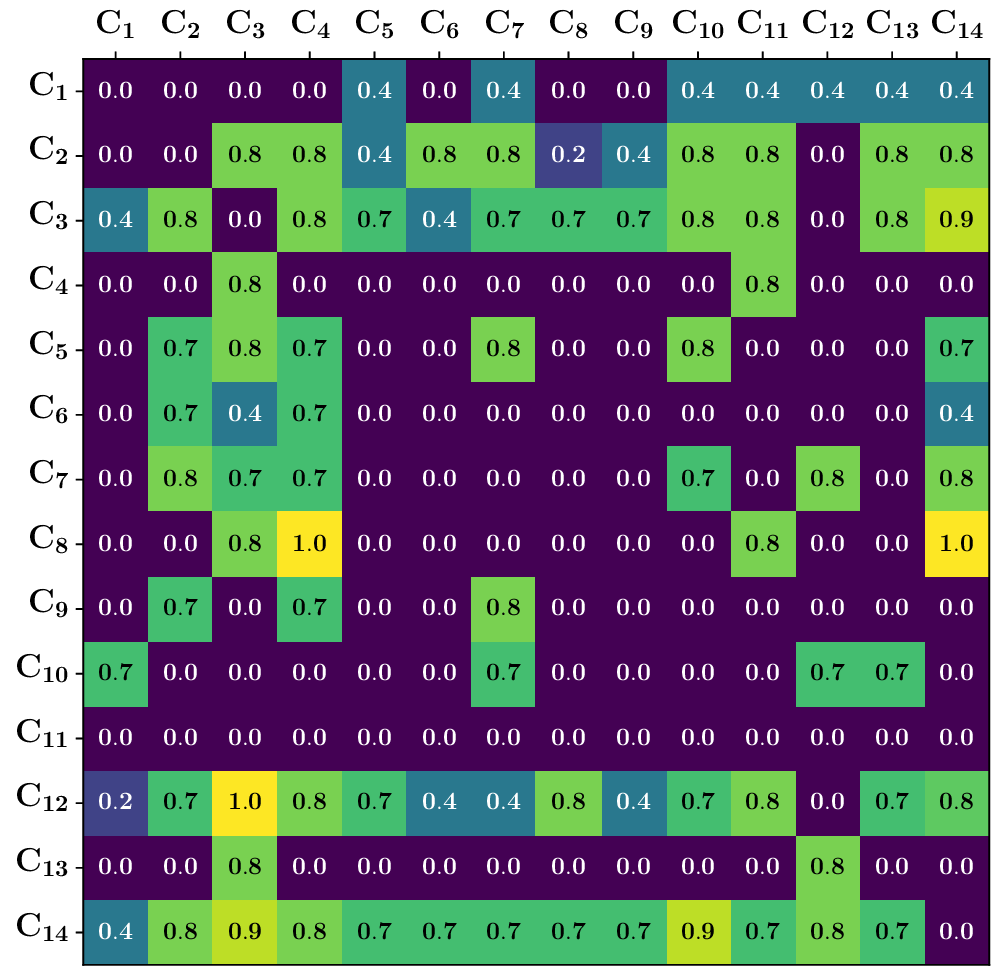}
\caption{\small{Adjusted edge-reconstruction ${\hat{E}}_2$ from latent II.}}
\label{fig:depression_prediction_refined_positive}
\end{subfigure}
\caption{Autoencoding for the clinical depression FCM: 
(a) Edge matrix $E$ corresponding to a 14-node FCM that models the causes of clinical depression. 
The nodes $C_1$--$C_{14}$ from the 1st column of Table~\ref{tab:depression_fcm} are along the rows and columns. 
The source nodes are along the rows and the target nodes are along the columns. 
The element on the ${i}^{\text{th}}$ row and ${j}^{\text{th}}$ column gives the weight on the edge from $C_i$ to $C_j$.  
Brighter color on the edge matrix corresponds to bigger edge weight or strength. 
(b) Reconstructed edge matrix $E_{1}$ from the encoded latent I summary.
The reconstruction is more accurate because the input text was detailed even if it sounded unnatural. 
(c) Reconstructed edge matrix $E_{2}$ from the encoded latent II summary that refined the encoded text with the content editing prompt or sub-task.
The reconstruction is not as accurate because the input text was not as detailed although it sounded natural. 
Many non-zero edge weights in $E$ changed to zero in ${E}_2$. 
The edge weights that correspond to the flipped nodes $C_4$, $C_9$, and $C_{10}$ from the $3^{\text{rd}}$ column of Table~\ref{tab:depression_fcm} are negative. 
These negative edges are in red. 
(d) The adjusted reconstructed edge matrix from latent II text with flipped nodes. 
The reconstruction is lossy but it preserves the stronger causal connections with larger edge weights. 
  }
  \vspace{-0.10in}
\label{fig:depression_fcm}
\end{figure*}

\begin{table*}[ht]
\caption{Target and reconstructed nodes for the autoencoding of the clinical depression FCM}
\begin{center}
\begin{tabular}{|c|l|l|l|}
    \hline
    \multirow{2}{*}{\bf Concept Node}&\multirow{2}{*}{\bf Target} & \multirow{2}{*}{\bf Reconstruction from Latent I } & \multirow{2}{*}{\bf Reconstruction from Latent II} \\
    & & & \\
    \hline
    $\mathbf{C_1}$ & Psychomotor agitation & Psychomotor agitation & Psychomotor agitation \\
    $\mathbf{C_2}$ & Psychomotor retardation & Psychomotor retardation & Psychomotor retardation  \\
    $\mathbf{C_3}$ & Depressive mood & Depressive mood & Depressive mood \\
    $\mathbf{C_4}$ & Reduced interest for daily function & Reduced interest for daily function & \textcolor{red}{Interest for daily function} \\
    $\mathbf{C_5}$ & Insomnia & Insomnia & Insomnia \\
    $\mathbf{C_6}$ & Hypersomnia & Hypersomnia & Hypersomnia \\
    $\mathbf{C_7}$ & Fatigue or loss of energy & Fatigue or loss of energy & Fatigue or loss of energy \\
    $\mathbf{C_8}$ & Recurrent thoughts of death & Recurrent thoughts of death & \textcolor{blue}{Thoughts of death} \\
    $\mathbf{C_9}$ & Loss of appetite & Loss of appetite & \textcolor{red}{Appetite} \\
    $\mathbf{C_{10}}$ & Diminished ability to think or concentrate & Diminished ability to think or concentrate & \textcolor{red}{Concentration} \\
    $\mathbf{C_{11}}$ & Indecisiveness & Indecisiveness & Indecisiveness \\
    $\mathbf{C_{12}}$ & Feelings of worthlessness & Feelings of worthlessness & \textcolor{blue}{Worthlessness}\\
    $\mathbf{C_{13}}$ & Extreme self-criticism & Extreme self-criticism & \textcolor{blue}{Self-criticism} \\
    $\mathbf{C_{14}}$ & Depression & Depression & Depression \\
    \hline
\end{tabular}
\vspace{-0.1in}
\end{center}
\label{tab:depression_fcm}
\end{table*}

\begin{table}[ht]
\caption{Reconstruction error with the autoencoding of the clinical depression FCM in Figure~\ref{fig:depression_target}}
\begin{center}
\begin{tabular}{|l|c|c|c|}
    \hline
    \multirow{2}{*}{\bf Metrics}&\multirow{2}{*}{\bf Latent I} & \multirow{2}{*}{\bf Latent II} & \multirow{2}{*}{\bf Adjusted from Latent II}  \\
    & & & \\
    \hline
    $\mathit{l}_{1}$-norm $\downarrow$ & {$\bf{14.56}$} & $78.40$ & $41.20$  \\
    $\mathit{l}_{2}$-norm  $\downarrow$ & $\bf{1.588}$ & $8.643$ & $4.240$ \\
    $\mathit{l}_{\infty}$-norm $\downarrow$ & $\bf{0.650}$ & $2.002$ & $0.653$ \\
    \hline 
\end{tabular}
\vspace{-0.2in} 
\end{center}
\label{tab:norm}
\end{table}

\begin{figure*}[!t]
\begin{subfigure}{0.32\linewidth}
\centering
\includegraphics[width=0.94\textwidth, height=0.94\textwidth]{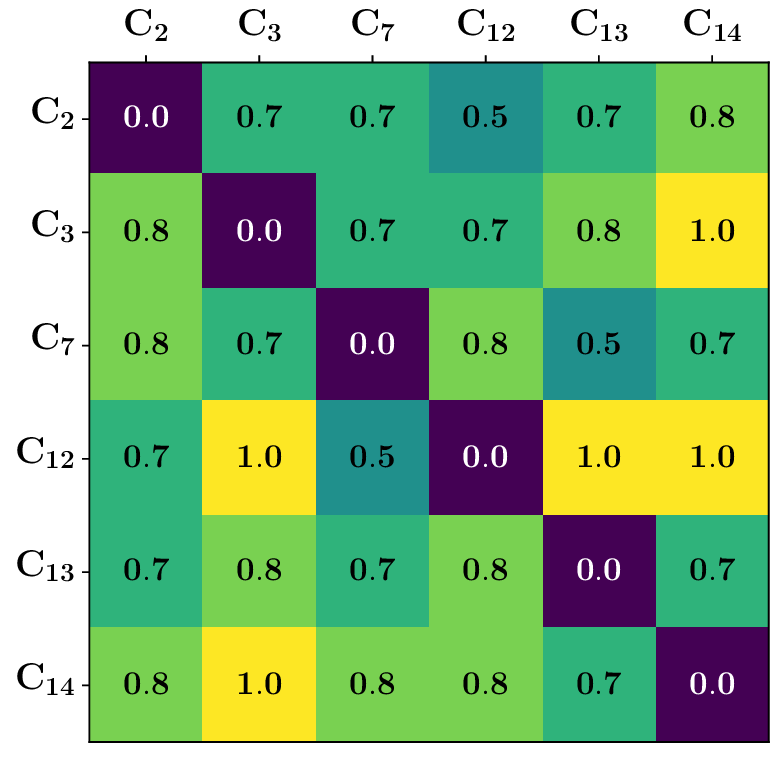}
\caption{\small{Target edge matrix: $E$}}
\label{fig:pruned_depression_target}
\end{subfigure}
\begin{subfigure}{0.32\linewidth}
\centering
\includegraphics[width=0.94\textwidth, height=0.94\textwidth]{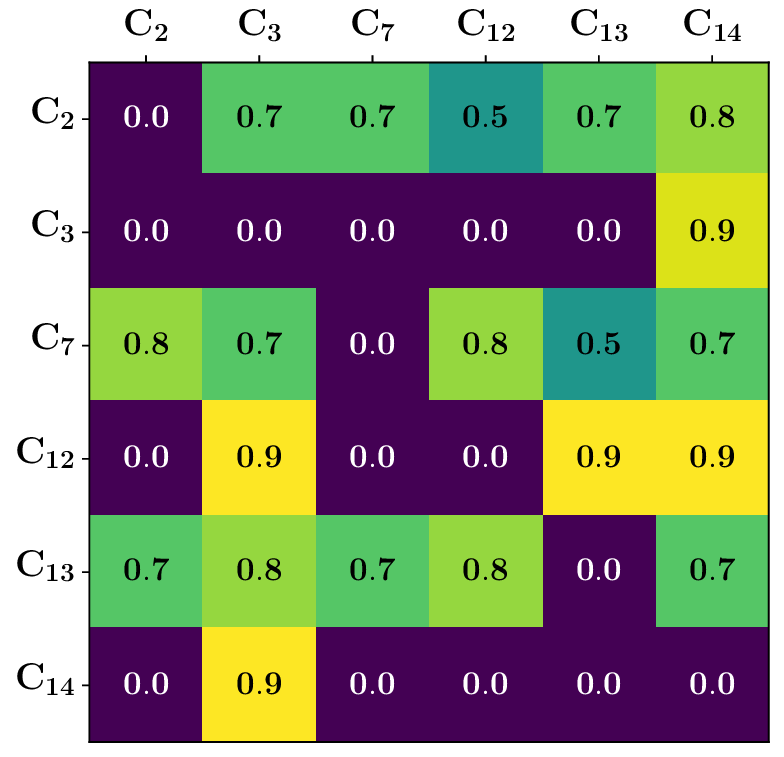} 
\caption{\small{Reconstructed edges from latent I: $E_1$}}
\label{fig:pruned_depression_prediction}
\end{subfigure}
\begin{subfigure}{0.32\linewidth}
\centering
\includegraphics[width=0.94\textwidth, height=0.94\textwidth]{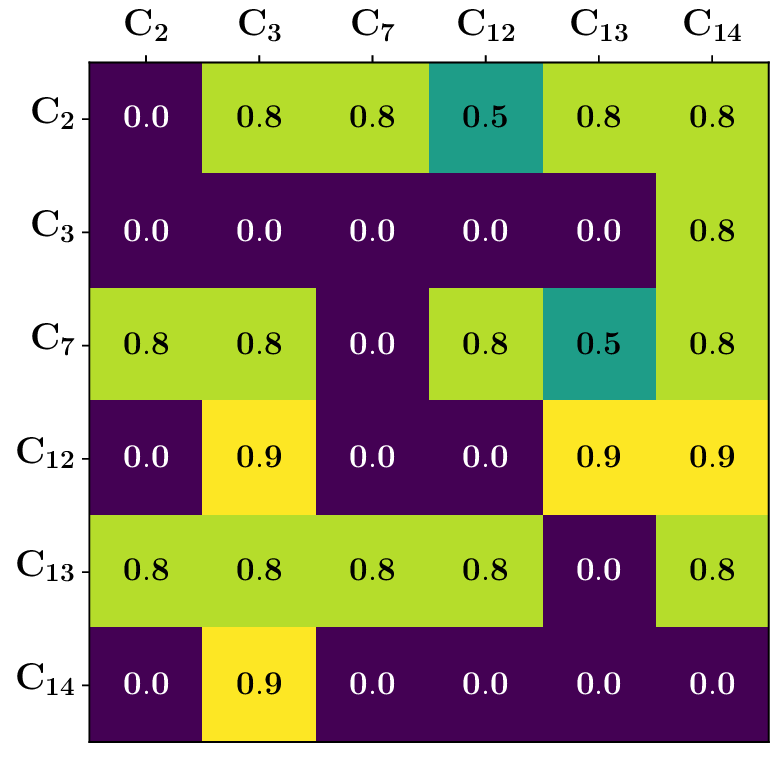}
\caption{\small{Reconstructed edges from latent II: $E_2$}}
\label{fig:pruned_depression_prediction_refined}
\end{subfigure}
\caption{FCM autoencoding for a strongly-connected depression FCM subset:
(a) The edge matrix $E$ corresponding to the subset of nodes $C_2$, $C_3$, $C_7$, $C_{12}$, $C_{13}$, and $C_{14}$ from the depression FCM model described by Table~\ref{tab:depression_fcm} and Figure~\ref{fig:depression_fcm}. 
The concept nodes from Table~\ref{tab:pruned_depression} index the rows and columns. 
The rows list the source nodes and the columns list the target nodes. 
The element on the ${i}^{\text{th}}$ row and the ${j}^{\text{th}}$ column gives the weight on the directed causal edge from the ${i}^{\text{th}}$ source node to the ${j}^{\text{th}}$ target node. 
The brighter colors correspond to the larger (stronger) causal edge weights. 
(b) Reconstructed edge matrix from the encoded latent I summary.
Many non-zero edge weights in $E$ are here zero but most of the bigger edge weights remain non-zero. 
(c) The reconstructed edge matrix from latent II summary that refined the encoded text with the content-editing prompt.
Many non-zero edge weights in $E$ changed to zero in $\hat{E}_2$  while most of the strongly connected edges remained non-zero.  
  }
\label{fig:pruned_depression_fcm}
\end{figure*}

\begin{table*}[ht]
\caption{Strongly-connected subset of the clinical depression FCM}
\begin{center}
\begin{tabular}{|c|l|l|l|}
    \hline
    \multirow{2}{*}{\bf Concept Node}&\multirow{2}{*}{\bf Target} & \multirow{2}{*}{\bf Reconstructed from Latent I } & \multirow{2}{*}{\bf Reconstructed from Latent II} \\
    & & & \\
    \hline
    $\mathbf{C_2}$ & Psychomotor retardation & Psychomotor retardation & Psychomotor retardation  \\
    $\mathbf{C_3}$ & Depressive mood & Depressive mood & Depressive mood \\
    $\mathbf{C_7}$ & Fatigue or loss of energy & Fatigue or loss of energy & Fatigue or loss of energy \\
    $\mathbf{C_{12}}$ & Feelings of worthlessness & Feelings of worthlessness & \textcolor{black}{Feelings of worthlessness}\\
    $\mathbf{C_{13}}$ & Extreme self-criticism & Extreme self-criticism & \textcolor{black}{Extreme self-criticism} \\
    $\mathbf{C_{14}}$ & Depression & Depression & Depression \\
    \hline
\end{tabular}
\end{center}
\label{tab:pruned_depression}
\end{table*}

\begin{figure*}[!t]
\begin{subfigure}{0.32\linewidth}
\centering
\includegraphics[width=0.94\textwidth, height=0.94\textwidth]{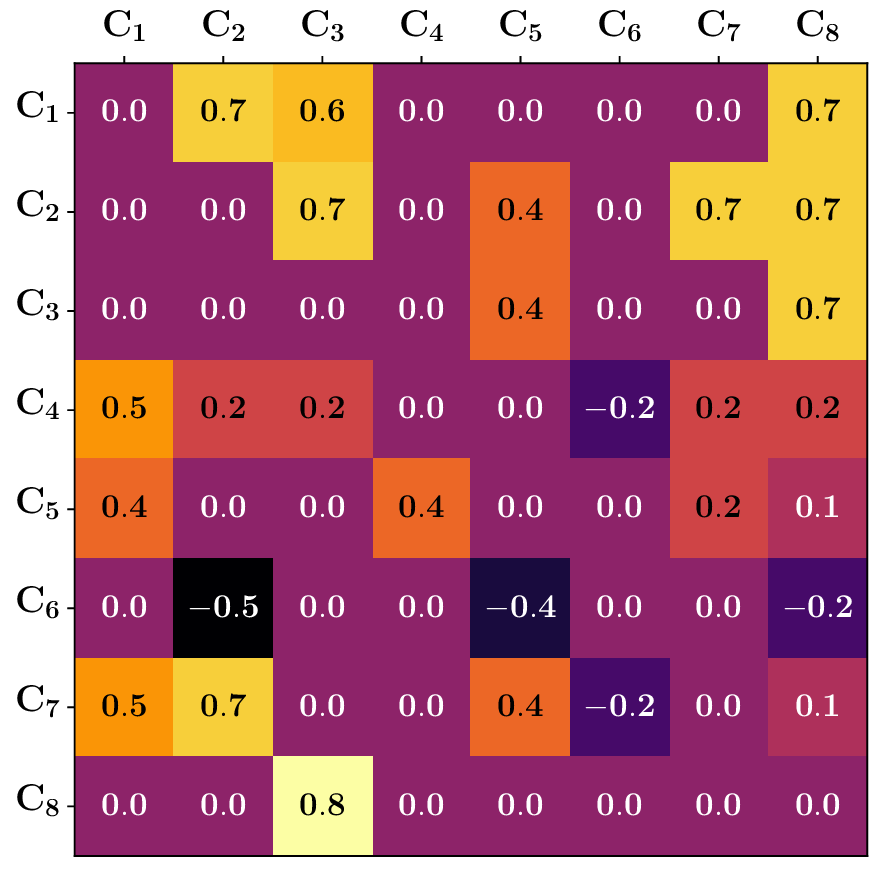}
\caption{\small{Target edge matrix: $E$}}
\label{fig:celiac_disease_target}
\end{subfigure}
\begin{subfigure}{0.32\linewidth}
\centering
\includegraphics[width=0.94\textwidth, height=0.94\textwidth]{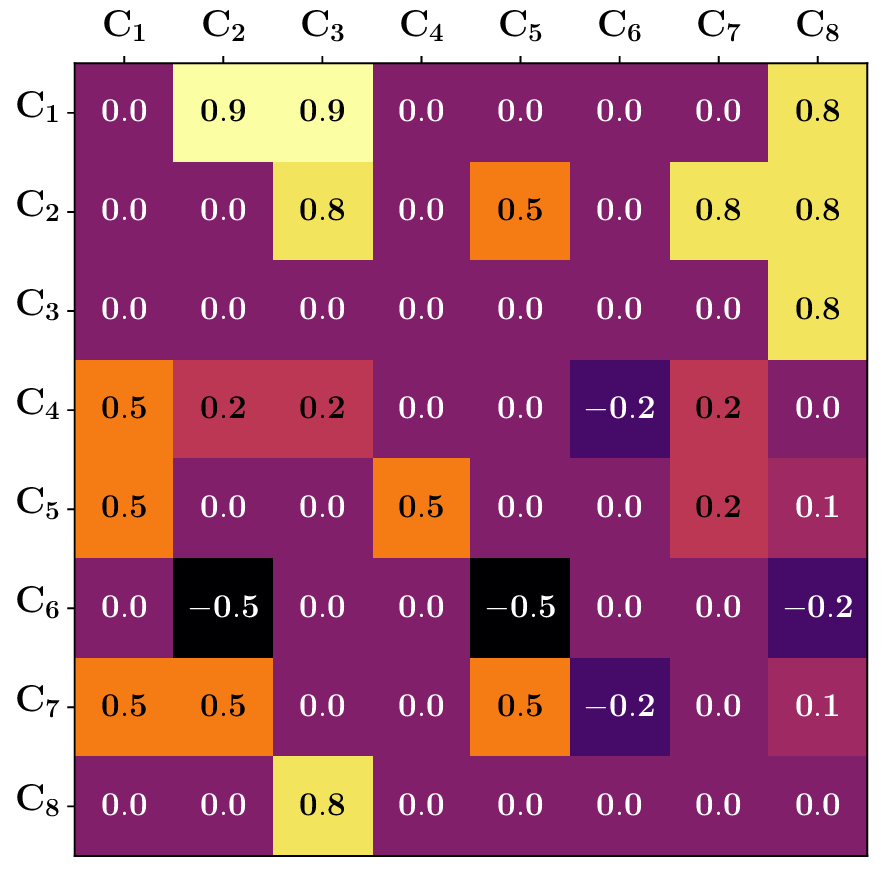} 
\caption{\small{Reconstructed edges from latent I: $E_1$}}
\label{fig:celiac_disease_prediction}
\end{subfigure}
\begin{subfigure}{0.32\linewidth}
\centering
\includegraphics[width=0.94\textwidth, height=0.94\textwidth]{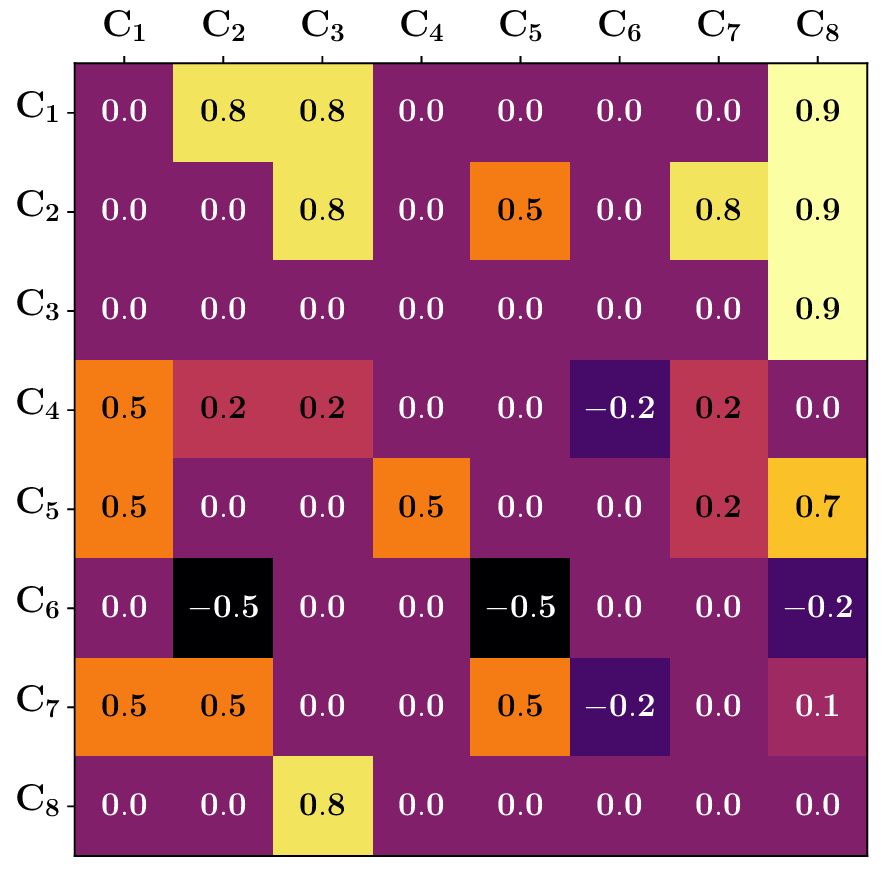}
\caption{\small{Reconstructed edges from latent II: $E_2$}}
\label{fig:celiac_disease_prediction_refined}
\end{subfigure}
\caption{FCM autoencoding for celiac disease classifier: 
(a) The edge matrix $E$ corresponding to the 8-node FCM model that classifies celiac disease. 
The concept nodes from table~\ref{tab:celiac_disease} index the rows and columns.  
The element on the ${i}^{\text{th}}$ row and the ${j}^{\text{th}}$ column gives the weight on the directed causal edge from $C_i$ to $C_j$. 
The brighter colors correspond to larger (stronger) causal edge weights. 
(b) Reconstructed edge matrix from the encoded latent I summary.
Many non-zero edge weights in $E$ are zero but most of the high-magnitude edge weights remain non-zero. 
(c) Reconstructed edge matrix from the encoded latent II summary that refined the encoded text with the content editing prompt or sub-task.
Many non-zero edge weights in $E$ are here zero but most of the high-magnitude edge weights remain non-zero.   
  }
\label{fig:celiac_disease}
\end{figure*}

\begin{table*}[ht]
\caption{Concept nodes of the Celiac Disease (CD) classifier FCM}
\begin{center}
\begin{tabular}{|c|l|l|l|}
    \hline
    \multirow{2}{*}{\bf Concept Node}&\multirow{2}{*}{\bf Target} & \multirow{2}{*}{\bf Reconstructed from Latent I} & \multirow{2}{*}{\bf Reconstructed from refined Latent II} \\
    & & & \\
    \hline
    $\mathbf{C_1}$ & Villi blunting & Villi blunting & \textcolor{blue}{Villous blunting}  \\
    $\mathbf{C_2}$ & Crypt hyperplasia & Crypt hyperplasia & Crypt hyperplasia \\
    $\mathbf{C_3}$ & Intraepithelial lymphocyte infiltration & Intraepithelial lymphocyte infiltration & Intraepithelial lymphocyte infiltration \\
    $\mathbf{C_{4}}$ & Epithelial changes & Epithelial changes & Epithelial changes \\
    $\mathbf{C_{5}}$ & Lamina propria MNC infiltration & Lamina propria MNC infiltration &  \textcolor{blue}{Lamina propria inflammation} \\
    $\mathbf{C_{6}}$ & Decrescendo pattern & Decrescendo pattern & Decrescendo pattern \\
    $\mathbf{C_{7}}$ & Mitoses & Mitoses & \textcolor{blue}{Mitotic activity} \\
    $\mathbf{C_{8}}$ & Class of celiac & Class of celiac & \textcolor{blue}{Classification of celiac} \\
    \hline
\end{tabular}
\vspace{-0.10in}
\end{center}
\label{tab:celiac_disease}
\end{table*}

\section{New FCM-LLM Mapping Technique}\label{Approach}

We designed a system with an LLM agent and multi-prompting to convert an input FCM to its latent summary and then reconstruct the FCM.
The system used successive system instructions together with multi-prompting.
The system instructions manipulate the behavior of the LLM agent. 
They tell the LLM agent how to process its inputs and how to structure its outputs. 
These instructions also specify what aspects of the input to focus on. 

\subsection{Encoding Prompt} 
This uses a set of system instructions that define how the LLM agent extracts the text summary {latent I} from the input FCM. 
This form of encoding maps the FCM to a detailed but unnatural summary.
It takes the FCM node list and the edge-weight matrix as input

The encoding prompt also instructs the LLM to explain each edge of the input FCM as text. 
The LLM summarizes the causal edge weights in words. 
The LLM has to measure the importance of each node based on how many edges connect it to other nodes. 
It then has to distribute the focus of the text among the nodes based on their importance: Focus more on the nodes that are more important. 
The LLM also has to sound natural as it describes this. 

\subsection{Content Editing Prompt} This prompt requires a set of system instructions that define how the LLM agent rewrites the text summary {latent II} from the encoded text summary output {latent I} of the LLM agent with the encoding prompt.  
The LLM agent with the encoding prompt may focus more describing the FCM edges than on sounding natural. 
It may generate text that sounds forced and repetitive and hard to read.
The LLM agent with the content editing prompt reworks the latent I summary and makes it sound more natural.
This system instruction is not too specific and depends on the LLM's  NLP capabilities. 

\subsection{Decoding Prompt}
The decoding divides into 3 subtasks: noun detection, node detection, and edge extraction.
The LLM agent uses a set of 3 successive system instructions to solve the subtasks.
\subsubsection{Noun detection} This system relies on the LLM's Named Entity Recognition (NER) capabilities. 
It asks the LLM to take the output text from the LLM agent with the encoding or the content editing prompt and process it sentence-by-sentence. 
The LLM then has to detect nouns, noun phrases, and pronouns in those sentences. 
The LLM also matches the pronouns to their corresponding noun antecedents. 
The nodes in an FCM are always nouns or noun phrases that describe a causal variable in the dynamical system. 
The detected nouns and noun phrases serve as node candidates for the reconstructed FCM. 
The LLM can also explain where these node candidates come from in the text. 
\subsubsection{Node detection} This system uses a set of system instructions to extract nodes from nouns and noun phrases.
It instructs the LLM agent to go through the list of nouns and noun phrases from node detection and then refines it into a list of FCM nodes. 
The FCM nodes are nouns and noun phrases that represent causal variables in the dynamical system. 
They have some degree of magnitude: They can either ``increase", ``improve", or ``intensify". 
The LLM agent with the `node detection' prompt goes through the list of nouns and noun phrases and looks for these properties. 
The LLM also checks if the text suggests a causal connection between the nouns/noun phrases. 
The LLM can quote from the text to give evidence of these causal connections. 
\subsubsection{Edge extraction} This system uses a set of instructions that describes how to extract $n^2-n$ node-pairs from a list of $n$ nodes. 
The LLM agent then goes through each node pair and looks through the text for evidence of positive, negative, or zero causal influence.
The LLM then assigns the edge weights based on the language used in the text. 
The LLM can also quote the text to justify its choice of causal edges. 

This list of edges along with the node-list completely describes the reconstructed FCM. 
The $2^{\text{nd}}$ column of Table~\ref{tab:depression_fcm} and Figure~\ref{fig:depression_prediction} give the respective nodes and the edge matrix $E_1$ of an FCM reconstructed from the LLM agent's ``raw" text summary latent I that described the FCM from Figure~\ref{fig:depression_target} and Table~\ref{tab:depression_fcm}'s column 1. 

The ${3}^{\text{rd}}$ column of Table~\ref{tab:depression_fcm} and Figure~\ref{fig:depression_prediction_refined} give the resepctive nodes and the edge matrix $E_2$ of the FCM reconstructed from the latent II text summary that describes the FCM from Figure~\ref{fig:depression_target} and Table~\ref{tab:depression_fcm}'s column 1.

\section{Experiment Setup}\label{Experiments}

We applied Google's  Gemini 2.5 Pro LLM with ``temperature = 0" and ``top\_p = 0.95" on three FCM inputs. 
The first FCM modeled the causes of clinical depression with 14 nodes and 180 edges. 
This densely connected FCM had only positive causal edges. 
The $1^{\text{st}}$ column of Table~\ref{tab:depression_fcm} lists its nodes $C_1$--$C_{14}$. 
Figure~\ref{fig:depression_target} shows its edge matrix $E$. 
The edge matrix $E$ uses brighter colors for causal edges with larger magnitudes. 

The LLM agent completely encoded the FCM into a latent I summary consisting of 824 words.
The LLM agent with the decoding prompt systematically reconstructed an FCM with 14 nodes but with 178 edges. 
The $2^{\text{nd}}$ column of Table~\ref{tab:depression_fcm} lists the 14 nodes $C_1$--$C_{14}$ and Figure~\ref{fig:depression_prediction} shows the edge matrix $E_1$. 
The edge matrix $E_1$ is also colored such that the brighter color corresponds to the higher edge weight.

The LLM agent with the content editing prompt reworked the LLM agent's 824-word latent I text into 602 words of refined latent II text. 
The LLM agent with decoding prompts reconstructed a 14-node FCM from this latent II text, but this FCM only had 89 edges. 
Also 27 out of those 89 edges were negative because 3 out of the 14 nodes represented the opposite of the corresponding target node. 
The $2^{\text{nd}}$ column of Table~\ref{tab:depression_fcm} lists the nodes $C_1$--$C_{14}$ of this FCM and colors the ``flipped" nodes in red. 
These nodes represent the opposite of their corresponding target node. 
Figure~\ref{fig:depression_prediction_refined} shows the edge matrix $E_2$ where brighter colors correspond to higher edge-weight magnitudes or causal-degree magnitudes. 
It colors the negative edge weights in red. 

The $2^{\text{nd}}$ FCM samples 6 nodes $C_2$, $C_3$, $C_7$, $C_{12}$, $C_{13}$, and $C_{14}$ from the first FCM. 
These nodes had the strongest causal connections among themselves. 
The $1^{\text{st}}$ column of Table~\ref{tab:pruned_depression} lists the nodes of this FCM and Figure~\ref{fig:pruned_depression_target} gives its edge matrix $E$. 
The figure colors the edge matrix $E$ so that brighter colors correspond to higher edge weights. 

The LLM agent describes this FCM in 211 words. 
The LLM agent with decoding prompts then reconstruct a 6-node FCM with 20 edges from this raw text. 
The $2^{\text{nd}}$ column of Table~\ref{tab:pruned_depression} lists the nodes of this FCM and Figure~\ref{fig:pruned_depression_prediction} shows the edge matrix from its latent I summary. 
The figure colors are such that the brighter colors correspond to larger edge weights. 

The LLM agent with the content editing prompt reworked the LLM agent's latent I summary into 321 words of refined latent II summary. 
The LLM agent with the decoding prompts reconstructed a 6-node FCM with 20 edges from this refined text. 
The $3^{\text{rd}}$ column of Table~\ref{tab:pruned_depression} and Figure~\ref{fig:pruned_depression_prediction_refined} show the respective nodes and the reconstructed edge matrix from the latent II summary. 
The figure colors are such that brighter colors correspond larger edge magnitudes. 

The $3^{\text{rd}}$ FCM describes classification of celiac disease (CD) from tissue using 8 nodes and 28 edges\cite{amirkhani2018novel}. 
The $1^{\text{st}}$ column of Table~\ref{tab:celiac_disease} and Figure~\ref{fig:celiac_disease_target} give the respective nodes and the edge matrix $E$ of this FCM. 
The LLM agent with the encoding prompt described this FCM with 211 words of the latent I summary. 
LLM agent with decoding prompts then reconstructed a 6-node FCM with 26 edges. 
The $2^{\text{nd}}$ column of Table~\ref{tab:celiac_disease} gives its nodes and Figure~\ref{fig:celiac_disease_prediction} gives its edges. 
The figure colors are also such that the brighter color corresponds to a higher edge weight.

The LLM agent with the content editor prompt reworked the output of the first FCM into 295 words of refined latent II text. 
The LLM agent with the decoding prompt reconstructed a 8-node 26-edge FCM out of this text. 
The $3^{\text{rd}}$ column of Table~\ref{tab:celiac_disease} and Figure~\ref{fig:celiac_disease_prediction_refined} give respective the nodes and the edge matrix $E_2$ of this FCM. 
The edge matrix shows the higher edge weights as brighter colors. 

\section{Discussion}\label{Discussion}

Figures~\ref{fig:depression_target} and \ref{fig:depression_prediction} show that the LLM agent with the encoding and the content editing prompt approximated an identity map from $E$ to $E_1$ pretty well despite never comparing $E_1$ to $E$. 
This is due to the careful step-by-step systematic method used to map the FCM to text and to map the text back to an FCM. 
The carefully designed system instructions multi-prompted the LLM agent to follow this method exactly. 

The latent I summary from the LLM agent with the encoding prompt sounded forced because the LLM focused more on accurately describing the FCM edges than sounding natural. 
The LLM agent with the content editing prompt reworded the latent I text summary to sound more natural but it came at the cost of reconstruction accuracy. 
Figures~\ref{fig:depression_target} and \ref{fig:depression_prediction_refined} show that the FCM reconstructed from the latent II summary of the LLM agent text missed a lot of edges. 
But even this lossy reconstruction preserved the stronger causal links of the FCM.  
The figures also show that edge weights on the $4^{\text{th}}$, $9^{\text{th}}$, and $10^{\text{th}}$ rows and columns have flipped signs. 
The edge weight $w_{49}$ flipped its sign twice and remained positive because it was in the $9^{\text{th}}$ row and the $4^{\text{th}}$ column. 
This occurred because the $4^{\text{th}}$, $9^{\text{th}}$, and $10^{\text{th}}$ reconstructed nodes from the latent II summary represented the opposite of the corresponding causal variable from the target FCM. 
Table~\ref{tab:depression_fcm}'s $1^{\text{st}}$ and $3^{\text{rd}}$ columns show this by highlighting the flipped nodes in red. 
The reconstructed FCM has ``appetite" as the $9^{\text{th}}$ node instead of ``loss of appetite" in the target FCM. 

Figure~\ref{fig:depression_prediction_refined_positive} flips the negative edges so we can compare it to \ref{fig:depression_target}. 
The comparison shows that many non-zero edges from Figure \ref{fig:depression_target} are zero in Figure \ref{fig:depression_prediction_refined_positive}. 
But most of the high-weight edges remained non-zero. 
So even the lossy reconstruction preserves most of the important edges. 

Table~\ref{tab:norm} measures the respective $l_1$, $l_2$, and $l_\infty$ norms of the reconstruction errors in its rows 1--3. 
Its $1^{\text{st}}$ and $2^{\text{nd}}$ columns measure the respective error involved in reconstructing the edge matrix from the latent I and latent II summaries.
The $3^{{rd}}$ column adjusts the reconstruction error from the latent II summary by flipping the edges connected to $C_4$, $C_9$, and $C_{10}$. 
The table shows that the latent I summary gave the best reconstructed edge matrix. 

Figures~\ref{fig:pruned_depression_target}--\ref{fig:pruned_depression_prediction_refined} and Figures~\ref{fig:celiac_disease_target}--\ref{fig:celiac_disease_prediction_refined} show something similar. 
Some non-zero edge weights from the target FCM are zero in the both the reconstructed FCMs from the latent I and the latent II summaries.
The two reconstructed FCMs differ little.
Both the reconstructed FCMs preserve the stronger causal connections from the target FCM. 

Tables~\ref{tab:depression_fcm} and \ref{tab:celiac_disease} show that there are some nodes in the FCM reconstructed from the latent II text that are slightly different from the corresponding target nodes. 
The tables highlight these nodes in blue. 
The $3^{\text{rd}}$ column of Table~\ref{tab:celiac_disease} calls the $7^{\text{th}}$ node ``mitotic activity" instead of ``mitoses" as in the $1^{\text{st}}$ column.

\section{Conclusion}\label{conclusion}

A sequence of well-designed system instructions can multi-prompt an LLM agent to convert an FCM into text and then reconstruct the FCM back from the text by a new multi-step method. 
This approximates an identity map from the FCM to itself like an autoencoder but with human-interpretable text descriptions and does not compare the reconstructed output FCM with the input FCM. 
The LLM agent also explains its decisions unlike black-box neural autoencoder models.

\bibliographystyle{unsrt}
\bibliography{bibdata}

\end{document}